\documentclass[10pt,twocolumn,letterpaper]{article}

\usepackage{wacv}
\usepackage{times}
\usepackage{epsfig}
\usepackage{amsmath}
\usepackage{amssymb}
\usepackage{booktabs}
\usepackage{subfigure}
\usepackage[ruled,vlined,linesnumbered]{algorithm2e}
\usepackage{color}
\usepackage{caption}
\usepackage{adjustbox}
\usepackage{multirow}
\usepackage{array}
\usepackage[font={small}]{caption}
\usepackage[pagebackref=true,breaklinks=true,letterpaper=true,colorlinks,bookmarks=false]{hyperref}

\wacvfinalcopy % *** Uncomment this line for the final submission

\def\wacvPaperID{358} % *** Enter the wacv Paper ID here

\def\ie{\textit{i.e.,~}}

\def\Vec#1{{\boldsymbol{#1}}}
\def\Mat#1{{\boldsymbol{#1}}}
\newtheorem{theorem}{Theorem}[section]

\newtheorem{proposition}[theorem]{Proposition}
\ifwacvfinal\fi
\begin{document}

%%%%%%%%% TITLE
\title{Automatic and Quantitative evaluation of attribute discovery methods}

% Authors at the same institution
%\author{First Author \hspace{2cm} Second Author \\
% Authors at different institutions
\author{Liangchen Liu\\
{\tt\small l.liu9@uq.edu.au}
\and
Arnold Wiliem\\
{\tt\small a.wiliem@uq.edu.au}
\and
Shaokang Chen\\
{\tt\small shaokangchenuq@gmail.com}
\and
Brian C. Lovell\\
{\tt\small lovell@itee.uq.edu.au}\\
The University of Queensland, School of ITEE\\
QLD 4072, Australia\\
}

\maketitle
\ifwacvfinal\thispagestyle{empty}\fi

%%%%%%%%% ABSTRACT
\begin{abstract}
	\label{abstract}
	
	Many automatic attribute discovery methods have been developed to extract a set of visual attributes from images for various tasks. 
	However, despite good performance in some image classification tasks, it is difficult to evaluate whether these methods discover meaningful attributes and which one is the best to find the attributes for image descriptions. 
	An intuitive way to evaluate this is to manually verify whether consistent identifiable visual concepts exist to distinguish between positive and negative images of an attribute. This manual checking is tedious, labor intensive and expensive
	and it is very hard to get quantitative comparisons between
	different methods. In this work, we tackle this problem by proposing an attribute meaningfulness metric, that can perform
	automatic evaluation on the meaningfulness of attribute sets as well as achieving quantitative comparisons. We apply our proposed metric to recent automatic
	attribute discovery methods and popular hashing methods
	on three attribute datasets. A user study is also conducted to
	validate the effectiveness of the metric. In our evaluation, we gleaned
	some insights that could be beneficial in developing automatic
	attribute discovery methods to generate meaningful
	attributes. To the best of our knowledge, this is the first
	work to quantitatively measure the semantic content of automatically
	discovered attributes.
	
\end{abstract}
\section{Introduction}
\label{sec_introduction}

\textit{``A picture is worth a thousand words''}.
This adage generally
refers to the notion that complex ideas can be explained
with a single picture. On the other hand, it can also be interpreted as, ``A thousand words are
required to explain a picture''. This has become one of the
emerging trends in the computer vision community~\cite{Farhadi09describingobjects,CHANGYLZH16,kumar2009attribute,feng2014learning,liu2014automatic,ChangYLZH2016}.
In this area, many attribute discovery methods have been developed to extract visual/image attributes for image description or classification~\cite{bergamo2011picodes,rastegari2012attribute,kovashka2015discovering,ChangYHXY15}.

One of the biggest challenges in using attribute descriptors
is that a set of labelled images is required to train the attribute classifiers. 
However, labelling each individual image for every attribute is a tedious, time-consuming
and expensive work especially when large number of images or attributes are required.  
Furthermore, in some specialized domains such as \textit{Ornithology}~\cite{WelinderEtal2010}, \textit{Entomology}~\cite{Wang2009learning} and cell
pathology~\cite{Wiliem2014discovering}, the labelling task could
be extremely expensive as only highly trained experts could do the
work.

\begin{figure}[!ht]
	\centering
	\includegraphics[width=1\linewidth]{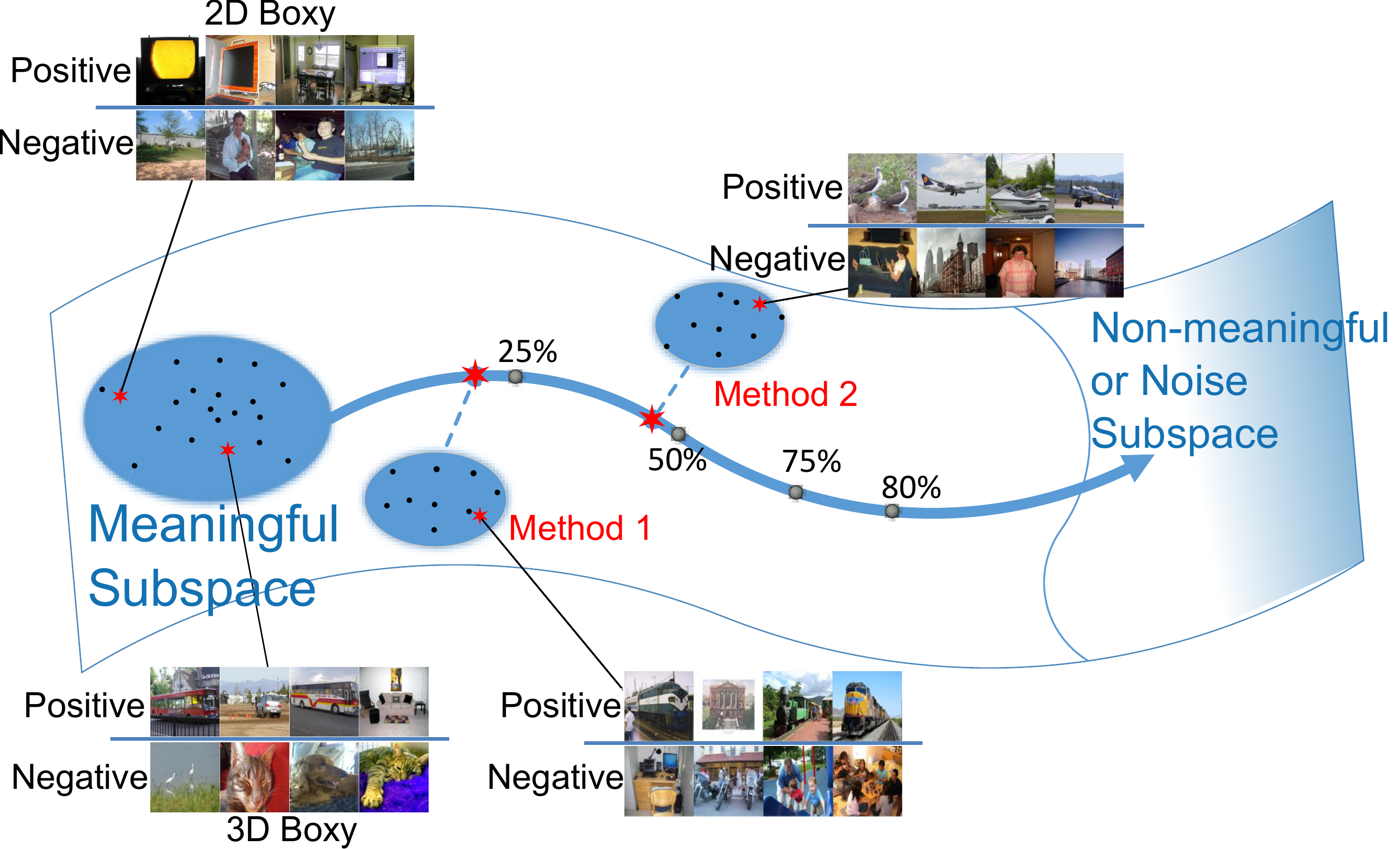}
	\caption{An illustration of the proposed attribute meaningfulness metric.
		Each individual attribute is represented as the outcome of the corresponding attribute classifier tested on a set of images.
		Inspired by~\cite{parikh2011interactively} we propose an approach to measure the distance between a set of discovered attributes and the meaningful subspace.
		The metric score is derived using a subspace interpolation between Meaningful Subspace and Non-meaningful/Noise Subspace.
		The score indicates how many meaningful attributes are contained in the set of discovered attributes.
	}
	\label{fig:idea}
\end{figure}

Therefore, automatic attribute discovery methods have been developed~\cite{bergamo2011picodes,rastegari2012attribute,sharmanska2012augmented,Wiliem2014discovering,yfx2013cvpr,ChangYYH15} for the labelling task. 
These works primarily focus on learning an embedding
function that maps the original descriptors into a binary code
space wherein each individual bit is expected to represent a visual attribute. 
We note that these approaches are also closely related to hashing methods~\cite{gong2011iterative,Leskovec2009Mining,weiss2009spectral}.
The difference is that unlike automatic attribute discovery approaches,
hashing methods are primarily aimed at significantly reducing computational complexity and storage whilst maintaining system accuracy.

Several problems that persist for the above methods are: 
1) Do these binary codes really have meaning? 
2) How do the codes extracted from one method compare with the others?
One of the benefits of studying these questions is we get to understand which
learning framework is required to develop an effective automatic
attribute discovery method that produces meaningful binary codes. 
To the best of our knowledge, this
is the first work to study these questions automatically and quantitatively. 
Note that, our focus is not to develop an automatic attribute discovery approach,
rather to compare various existing automatic attribute discovery
approaches and determine the ones that consistently discover
meaningful attributes.

Gauging ``how meaningful'' for a given attribute could be an
ill-posed problem, as there is not any \textit{yardstick} which
could be used to measure this. 
In recent works, Parikh and Grauman
speculated that there is \textit{shared structure} among
meaningful attributes~\cite{parikh2011interactive,parikh2011interactively}. 
It is assumed that meaningful attributes
are close to each other. In~\cite{Liu2016isba}, an automatic keywords generation approach is proposed
for describing surveillance video based on this assumption. 
Inspired by these researches,
we propose a novel metric to become one of the yardsticks for measuring attribute
meaningfulness. 
More specifically, we define the distance between an attribute set and a meaningful
attribute subspace by measuring reconstruction errors to estimate the meaningfulness. 
Then a metric is derived by subspace interpolation between the meaningful subspace and the non-meaningful subspace to calibrate the distance. 
The metric can quantitatively determine how much meaningful content is contained in a set of attributes discovered automatically.
Fig.~\ref{fig:idea} illustrates our key ideas.

%To make easy interpretation on the metric, we utilise the geodesic between the Meaningful Subspace to Non-Meaningful Subspace.
%Using this, the metric score can be interpreted as ``how much noise/non-meaningful attributes contained in the set of attributes''.
%To our knowledge, this is the first study to measure the semantic content of a set of discovered attributes.
\noindent
\textbf{Contributions --- } We list our contributions as follows:
\begin{itemize}
	\vspace{-1.5ex}
	\item  We propose a reconstruction error based approach with $\ell_0$ and convex hull regularizers to approximate the distance of a given attribute set from the Meaningful Subspace.
	\vspace{-1.5ex}
	\item We propose the \textit{attribute meaningfulness} metric that
	allows us to quantitatively  measure the meaningfulness of a set of attributes.
	The metric score is related to ``the percentage of meaningful attributes contained in the set of attributes''.
	\vspace{-1.5ex}
	\item We show in the experiments that our proposed metric indeed captures the meaningfulness of attributes. We also study the \textit{attribute meaningfulness} of some recent automatic attribute discovery
	methods and various hashing approaches on three
	attribute datasets.
	A user study is also applied on two datasets to further show the effectiveness of the proposed metric. 
\end{itemize}

\vspace{-1ex}
We continue our paper as follows.
The related work is discussed in Section~\ref{sec_related}.
We then introduce our approach to measure attribute meaningfulness in Section~\ref{sec:measuring_meaningfulness}.
Our proposed metric is described in Section~\ref{sec:part3}.
The experiments and results are discussed in Section~\ref{Experiment}.
Finally the main findings and future directions are presented in Section~\ref{sec_conclusion}.

\section{Related Work}
\label{sec_related}

Traditional evaluation of visual attribute meaningfulness are done manually by observing whether consistent identifiable visual concepts are present/absent in a set of given images. 
Generally, crowd-sourcing systems such as the Amazon Mechanical Turk~(AMT)~\footnote{www.mturk.com} are used for this task. 
However, this is ineffective and expensive because we need to repeat this process whenever new attributes are generated or novel methods are proposed.
For instance, the AMT Human Intelligence Task (HIT) for our case is to evaluate the meaningfulness of attributes by examining the corresponding positive and negative images. 
Assuming it requires two minutes on average to verify one attribute, an AMT worker may require 
320 minutes for evaluating 32
attributes discovered by 5 different methods (\ie $32 \times 5
\times 2 = 320$ minutes). 
This could increase significantly when
more AMT workers are required to produce statistically reliable results.

%,our method only needs 5 secs.
A more cost-effective, less labor intensive and time consuming method is to develop an automatic approach to evaluate the meaningfulness of the set of discovered attributes. 
In their work, Parikh and Grauman~\cite{parikh2011interactively} assumed that there is shared structured among meaningful attributes. 
They proposed an active learning approach that uses probabilistic principal component analyzers (MPPCA)~\cite{tipping1999mixtures}
to predict how likely the visual division created by an attribute is nameable. 
Note that, when an attribute is nameable, it is assumed to be meaningful. 
Nevertheless, it is not clear how to use their approach to perform a quantitative measurement, as the predictor only decides whether an attribute is nameable or not.
In addition, their method is semi-automatic as human intervention is required in the process. 
Thus, their method is not suitable for addressing our goal 
(i.e., to automatically evaluate the meaningfulness of attribute sets).

In~\cite{Liu2016isba}, a selection method is proposed for attribute discovery methods to assist attribute-based keywords generation for video description from surveillance systems. Their selection method is based on the shared structure assumption. However, this work did not consider quantitative analysis of the meaningfulness of the discovered attributes (\eg how much meaningful content contained is in a set of attributes). Moreover distances proposed in~\cite{Liu2016isba} might not be sufficient to capture all the characteristics of the meaningfulness of attributes as it does not reflect the direct correlation between each meaningful attribute with the discovered ones.

\section{Measuring Attribute Set Meaningfulness}
\label{sec:measuring_meaningfulness}
%writing guide 
%This meaningfulness manifold

%the concept of distance has also been described in cite that paper (in the ref)

%use same symbols the same 

%maybe another story  

In this section, we first introduce the manifold of decision boundaries and meaningful attribute subspace. 
Then, we define the distance between automatically discovered attributes and the meaningful attribute set in this space to measure the attribute meaningfulness.

\subsection{Manifold of decision boundaries}
\label{sec:manifold}

Given a set of $N$ images, $\mathcal{X} = \{ \Vec{x}_i \}_{i=1}^N$, an attribute can be considered as a decision boundary which divides the set into two disjoint subsets $\mathcal{X}^+ \cup \mathcal{X}^- = \mathcal{X}$, where $\mathcal{X}^{+}$ represent the set of images where the attribute exists and $\mathcal{X}^{-}$ represents the set of images where the attribute is absent.
Thus, all the attributes are lying on a manifold formed by decision boundaries~\cite{parikh2011interactively,semi-MA}.

An attribute can also be viewed as a binary vector of $N$-dimensions.
The $i$-th element of the binary vector represents the output of sample $\Vec{x}_i$ tested by the corresponding attribute binary classifier denoted as $\phi ( \cdot ) \in \mathbb{R}$.
The sign of the classifier output on $\Vec{x}_i$ indicates whether the sample belongs to the positive or negative set (\ie $\mathcal{X}^+$ or $\mathcal{X}^-$). 
Given a set of samples $\mathcal{X}$, attribute representation is defined as $\Vec{z}^{[\mathcal{X}]} \in \{-1,+1\}^{N}$ whose the $i$-th element is $\Vec{z}_{(i)}^{[\mathcal{X}]} = \operatorname{sign}(\phi(\Vec{x}_i)) \in \{-1,+1\}$.
For simplicity of symbols we use $\Vec{z}$ instead of $\Vec{z}^{[\mathcal{X}]}$ here.

As such, we define the manifold of decision boundaries w.r.t. $\mathcal{X}$ as 
$\mathcal{M}^{[\mathcal{X}]} \in \{-1,+1\}^{N}$ which is embedded in a $N$-dimensional binary space.

As observed from [13, 14], the meaningful attributes share the same structure that lie close to each other on the manifold. That is, all the meaningful attributes form a subspace in $\mathcal{M}^{[\mathcal{X}]}$ within a limited region. Ideally, this subspace should contain all possible meaningful attributes. 
Unfortunately, in practical cases, it is infeasible to enumerate
all of these. 
One intuitive solution is to represent the meaningful subspace by using human labelled attributes from various image
datasets such as \cite{biswas2013simultaneous,parikh2011interactive,parikh2011interactively}. 
As they are annotated by
human annotators via AMT, they are all naturally meaningful. 
Unequivocally, human labelled attributes are very limited in terms of the number and their meanings. 
We thus introduce an approximation of the meaningful subspace by linear combinations of the human labelled attributes. 
That is, if an attribute is close to any attribute lying in the meaningful subspace, it is considered to be a meaningful attribute. 

\subsection{Distance of an attribute to the Meaningful Subspace}
\label{sec:distance}

Here our goal is to define the distance of an attribute from the Meaningful Subspace. 
Given a set of $N$ images $\mathcal{X}$, we use $\mathcal{S} = \{\Vec{h}_j\}_{j=1}^J, \Vec{h}_j \in \{-1,+1\}^N$ to denote the set of meaningful attributes.
%Then we introduce matrix $\Mat{A} \in \mathbb{R}^{N \times J}$ to represent the set $\mathcal{S}$ with the attributes in $\mathcal{S}$ as its column vector, given we have $J$ attributes.
Let $\Mat{A} \in \mathbb{R}^{N \times J}$ be a matrix in which each column vector is the representation of a meaningful attribute.
According to~\cite{parikh2011interactively}, meaningful attributes should be close to the meaningful subspace $\mathcal{S}$. 
The meaningful subspace is spanned by the set of meaningful attributes.
For example, the set of secondary colors such as yellow, magenta and cyan, can be reconstructed from primary colors \ie red, green, blue.  
Even sometimes the primary colors can provide clues to describe other primary colors with negative information~(\eg red is not green and not blue).
With this in mind, we can define the distance between an attribute and the meaningful subspace via its reconstruction error.
More specifically, let $\Vec{z}_k$ be an attribute and $\Mat{A}$ is the meaningful attribute representation. The distance is defined as:
\begin{equation}
\min_\Vec{r} \| \Mat{A} \Vec{r}  - \Vec{z}_k \|_2^2,
\label{eq:single_dist}
\end{equation}
\noindent
where $\Vec{r} \in \mathbb{R}^{J \times 1}$ be the reconstruction coefficient vector.
Note that the reconstruction expressed in the above equation does not always lie on the manifold $\mathcal{M}$. 
Therefore, the distance is considered as a first-order approximation.

\subsection{Distance between a set of discovered attributes and the Meaningful Subspace}
Similarly, we introduce matrix $\Mat{B} \in \mathbb{R}^{N \times K}$ to represent the discovered attribute set $\mathcal{D}$ which contains $K$ discovered attributes.
We then define the distance between the set of discovered attributes $\mathcal{D}$ and the Meaningful Subspace $\mathcal{S}$ w.r.t. the set of images $\mathcal{X}$ as the average reconstruction error:
\begin{equation}
\delta(\mathcal{D},{\mathcal{S}}; \mathcal{X}) = \frac{1}{K}\min_{\Mat{R}} \| \Mat{A}\Mat{R} - \Mat{B} \|_F^2,
\label{eq:min_dist}
\end{equation}
where $\| \cdot \|_F$ is the matrix Frobenious norm;
$\Mat{R} \in \mathbb{R}^{J \times K}$ is the reconstruction matrix.

The distances in~\eqref{eq:single_dist} and~\eqref{eq:min_dist} may create dense 
reconstruction coefficients, suggesting that each meaningful attribute 
should contribute to the reconstruction.
A more desired result is to have less dense coefficients (\ie fewer non-zero coefficients).
This is because there may be only a few meaningful attributes required 
to reconstruct another meaningful attribute.
As such, we consider convex hull regularization first used in~\cite{Liu2016isba}.
Moreover, the perception characteristics of human visual systems favor sparse responses~\cite{serre2007robust}.
That means only a few interesting obvious attributes will first trigger the semantic-visual connection in our brain.
Attribute discovery methods should also follow this procedure. Accordingly, we propose the second regularization, the $\ell_0$ regularization.

\vspace{-1ex}
\subsubsection{Convex hull regularization}

When a convex hull constraint is considered,~\eqref{eq:min_dist} becomes:
\begin{equation}
\label{eq:con_hull}
\begin{aligned}
\delta_{\operatorname{cvx}}(\mathcal{D},\mathcal{S}; \mathcal{X}) = \frac{1}{K}\min_{\Mat{R}} \| \Mat{A}\Mat{R} - \Mat{B} \|_F^2   \operatorname{s.t.} \\
{\Mat{R}}\left( {i,j} \right) \ge 0 & \\
\sum_{i=1}^{J} {\Mat{R}(i,\cdot)}=1. &
\end{aligned}	
\end{equation}
\noindent
The above equation basically computes the average distance between each discovered attribute $\Vec{z}_k \in \mathcal{D}$ and the convex hull of $\mathcal{S}$.
The above optimization problem can be solved using the method proposed in~\cite{cevikalp2010face}.
%We use the same technique to solve the above optimization problem as in \cite{cevikalp2010face}. 

\subsubsection{$\ell_0$ regularization}

Unlike the convex hull regularization, here we consider a possible 
direct correlation between each discovered attribute $\Vec{z}_k \in \mathcal{D}$ 
and the meaningful attribute, $\Vec{h}_j \in \mathcal{S}$:
\begin{align}	
\label{eq:joint_prob}
\delta_{\operatorname{jp}}(\mathcal{D},{\mathcal{S}}; \mathcal{X}) = & \frac{1}{K} \min_{\Mat{R}} \| \Mat{A}\Mat{R} - \Mat{B} \|_F^2, \operatorname{s.t.} \\
& \forall k \in \{ 1 \cdots K \}, \| \Mat{R}_{\cdot,k} \|_0 \leq 1, \nonumber \\
& \forall j \in \{ 1 \cdots J \}, \| \Mat{R}_{j,\cdot} \|_0 \leq 1. \nonumber	
\end{align}			
\noindent
where $\Mat{R}_{j,\cdot}$ is the row vector of the $j$-th row of
matrix $\Mat{R}$ and $\Mat{R}_{\cdot,k}$ is the column vector of 
the $k$-th column of matrix $\Mat{R}$.
These additional constraints enforce one-to-one relationships between ${\mathcal{S}}$ and $\mathcal{D}$.
The matrix $\Mat{R}$ relates individual discovered attributes to each meaningful attribute.
In other words, for each discovered attribute $\Vec{z}_k \in \mathcal{D}$,
we would like to find the closest $\Vec{h}_j \in \mathcal{S}$ that
minimizes the above objective.
Note that, when $|{\mathcal{S}}| > |\mathcal{D}|$ then we can only match $K$ discovered attributes in ${\mathcal{S}}$ and vice versa.

Unfortunately, these constraints make the optimization problem~\eqref{eq:joint_prob} non-convex.
Thus, we propose a greedy approach to address this
by finding pairs of meaningful and discovered attributes that 
have the lowest distance.
This is equivalent to finding the pairs that have the the highest similarities (lowest distance means high similarity).
%We can interpret the problem~\eqref{eq:joint_prob} as finding the 
%pairs of meaningful and discovered 
%attributes that have highest similarities.
%To see this, let us consider the extreme case where $\mathcal{D} = \mathcal{S}$.
%Here, both have the highest similarities (in fact both are the same).
%In addition, as the matrix $\Mat{R}$ becomes the identity matrix, the function $\delta_{\operatorname{jp}}$ is minimized (\ie $= 0$).

The similarities between a meaningful attribute $\Vec{h}_j$  
and a discovered attribute $\Vec{z}_k$ can 
be defined in terms of their correlations.
Let $\rho(\Vec{z}_k, \Vec{h}_j), \Vec{z}_k \in \mathcal{D}, \Vec{h}_j \in \mathcal{S}$ be the correlation between $\Vec{z}_k$ and $\Vec{h}_j$. 
The correlation $\rho$ is defined via:
\begin{equation}
\rho(\Vec{z}_k, \Vec{h}_j) = \frac{\operatorname{count} (\Vec{z}_k = \Vec{h}_j)}{N},
\end{equation}
\noindent
where $\operatorname{count}$ is a function that counts the number of times elements in $\Vec{z}_k$ equal elements in $\Vec{h}_j$.

The function $\rho(\Vec{z}_k, \Vec{h}_j)$ can be determined from $\Mat{A}_{\cdot, j}$ and $\Mat{B}_{\cdot,k}$, where 
$\Mat{A}_{\cdot, j}$ is the meaningful attribute $\Vec{h}_j$ 
and $\Mat{B}_{\cdot,k}$ is the discovered attribute $\Vec{z}_k$.
Let $\mathcal{P}$ be the set of M pairs of $\Vec{h}_j \in \mathcal{S}$ and $\Vec{z}_k \in \mathcal{D}$ that have the highest correlation, 
$\mathcal{P} = \{ ( \Vec{h}^1_j, \Vec{z}^1_k ) \cdots ( \Vec{h}^M_j, \Vec{z}^M_k )\}$, 
$\Vec{h}^i_j = \Vec{h}^l_j$ if and only if $i = l$, $\Vec{z}^i_k = \Vec{z}^l_k$ if and only if $i = l$.

Once $\mathcal{P}$ is determined, the matrix $\Mat{R}^*$ that minimizes~\eqref{eq:joint_prob} is defined via:
\begin{equation}
\label{eq:jp_R}	
\Mat{R}^*_{j,k} = \left\{ \begin{array}{l}\ 1\ \operatorname{if}\ ( \Vec{h}_j,\Vec{z}_k)  \in  {\cal P}\\
\ 0\ \operatorname{if}\ ( \Vec{h}_j,\Vec{z}_k)  \notin  {\cal P}.
\end{array} \right.
\end{equation}

Algorithm~\ref{algo:joint_prob} computes the set $\mathcal{P}$ for given input $\mathcal{D} = \{ \Vec{z}_k \}_{k=1}^K$, ${\mathcal{S}} = \{ \Vec{h}_j \}_{j=1}^{J}$  and $\mathcal{X} = \{ \Vec{x}_i \}_{i=1}^N$.
Note that, $( \Vec{h}_j, \cdot )$ and $( \cdot, \Vec{z}_k )$, used in step 3, represent all possible pairs that contain $\Vec{h}_j$ and $\Vec{z}_k$, respectively.

\subsubsection{$\ell_0$ regularization}

Unlike the convex hull regularization, here we consider a possible 
direct correlation between each discovered attribute $\Vec{z}_k \in \mathcal{D}$ 
and the meaningful attribute, $\Vec{h}_j \in \mathcal{S}$:
\begin{align}	
\label{eq:joint_prob}
\delta_{\operatorname{jp}}(\mathcal{D},{\mathcal{S}}; \mathcal{X}) = & \frac{1}{K} \min_{\Mat{R}} \| \Mat{A}\Mat{R} - \Mat{B} \|_F^2, \operatorname{s.t.} \\
& \forall k \in \{ 1 \cdots K \}, \| \Mat{R}_{\cdot,k} \|_0 \leq 1, \nonumber \\
& \forall j \in \{ 1 \cdots J \}, \| \Mat{R}_{j,\cdot} \|_0 \leq 1. \nonumber	
\end{align}			
\noindent
where $\Mat{R}_{j,\cdot}$ is the row vector of the $j$-th row of
matrix $\Mat{R}$ and $\Mat{R}_{\cdot,k}$ is the column vector of 
the $k$-th column of matrix $\Mat{R}$.
These additional constraints enforce one-to-one relationships between ${\mathcal{S}}$ and $\mathcal{D}$.
The matrix $\Mat{R}$ relates individual discovered attributes to each meaningful attribute.
In other words, for each discovered attribute $\Vec{z}_k \in \mathcal{D}$,
we would like to find the closest $\Vec{h}_j \in \mathcal{S}$ that
minimizes the above objective.
Note that, when $|{\mathcal{S}}| > |\mathcal{D}|$ then we can only match $K$ discovered attributes in ${\mathcal{S}}$ and vice versa.

Unfortunately, these constraints make the optimization problem~\eqref{eq:joint_prob} non-convex.
Thus, we propose a greedy approach to address this
by finding pairs of meaningful and discovered attributes that 
have the lowest distance.
This is equivalent to finding the pairs that have the the highest similarities (lowest distance means high similarity).
%We can interpret the problem~\eqref{eq:joint_prob} as finding the 
%pairs of meaningful and discovered 
%attributes that have highest similarities.
%To see this, let us consider the extreme case where $\mathcal{D} = \mathcal{S}$.
%Here, both have the highest similarities (in fact both are the same).
%In addition, as the matrix $\Mat{R}$ becomes the identity matrix, the function $\delta_{\operatorname{jp}}$ is minimized (\ie $= 0$).

The similarities between a meaningful attribute $\Vec{h}_j$  
and a discovered attribute $\Vec{z}_k$ can 
be defined in terms of their correlations.
Let $\rho(\Vec{z}_k, \Vec{h}_j), \Vec{z}_k \in \mathcal{D}, \Vec{h}_j \in \mathcal{S}$ be the correlation between $\Vec{z}_k$ and $\Vec{h}_j$. 
The correlation $\rho$ is defined via:
\begin{equation}
\rho(\Vec{z}_k, \Vec{h}_j) = \frac{\operatorname{count} (\Vec{z}_k = \Vec{h}_j)}{N},
\end{equation}
\noindent
where $\operatorname{count}$ is a function that counts the number of times elements in $\Vec{z}_k$ equal elements in $\Vec{h}_j$.

The function $\rho(\Vec{z}_k, \Vec{h}_j)$ can be determined from $\Mat{A}_{\cdot, j}$ and $\Mat{B}_{\cdot,k}$, where 
$\Mat{A}_{\cdot, j}$ is the meaningful attribute $\Vec{h}_j$ 
and $\Mat{B}_{\cdot,k}$ is the discovered attribute $\Vec{z}_k$.
Let $\mathcal{P}$ be the set of M pairs of $\Vec{h}_j \in \mathcal{S}$ and $\Vec{z}_k \in \mathcal{D}$ that have the highest correlation, 
$\mathcal{P} = \{ ( \Vec{h}^1_j, \Vec{z}^1_k ) \cdots ( \Vec{h}^M_j, \Vec{z}^M_k )\}$, 
$\Vec{h}^i_j = \Vec{h}^l_j$ if and only if $i = l$, $\Vec{z}^i_k = \Vec{z}^l_k$ if and only if $i = l$.

Once $\mathcal{P}$ is determined, the matrix $\Mat{R}^*$ that minimizes~\eqref{eq:joint_prob} is defined via:
\begin{equation}
\label{eq:jp_R}	
\Mat{R}^*_{j,k} = \left\{ \begin{array}{l}\ 1\ \operatorname{if}\ ( \Vec{h}_j,\Vec{z}_k)  \in  {\cal P}\\
\ 0\ \operatorname{if}\ ( \Vec{h}_j,\Vec{z}_k)  \notin  {\cal P}.
\end{array} \right.
\end{equation}

Algorithm~\ref{algo:joint_prob} computes the set $\mathcal{P}$ for given input $\mathcal{D} = \{ \Vec{z}_k \}_{k=1}^K$, ${\mathcal{S}} = \{ \Vec{h}_j \}_{j=1}^{J}$  and $\mathcal{X} = \{ \Vec{x}_i \}_{i=1}^N$.
Note that, $( \Vec{h}_j, \cdot )$ and $( \cdot, \Vec{z}_k )$, used in step 3, represent all possible pairs that contain $\Vec{h}_j$ and $\Vec{z}_k$, respectively.

\begin{algorithm}[t]
	\caption{The proposed greedy algorithm to solve~\eqref{eq:joint_prob}}
	\label{algo:joint_prob}
	Input: $\mathcal{D} = \{ \Vec{z}_k \}_{k=1}^K$, ${\mathcal{S}} = \{ \Vec{h}_j \}_{s=1}^{J}$  and $\mathcal{X} = \{ \Vec{x}_i \}_{i=1}^N$.
	Output: $\mathcal{P}$ that contains M pairs that have the highest correlation, where $M = \min ( K, J )$.
	
	$\mathcal{P} \leftarrow \{ \}$.
	
	\Repeat{$|\mathcal{P}| \le M$}{
		Find the highest $\rho(\Vec{h}_j, \Vec{z}_k)$ where $( \Vec{h}_j, \cdot ) \notin \mathcal{P}$ and $( \cdot, \Vec{z}_k ) \notin \mathcal{P}$
		
		$\mathcal{P} = \mathcal{P} \cup (\Vec{h}_j, \Vec{z}_k )$
		}
\end{algorithm}

\section{Attribute Set Meaningfulness Metric}
\label{sec:part3}

%Just as figure~\ref{fig1:idea}, granted a manifold space exists for the visual attributes, there is a subspace accounting for the meaningful ones with shared structure within themselves.
%The automatic discovery methods A and B \etc. can be measured by "ruler" starting from the subspace of meaningful visual attribute to the completely random ones. 

The distance functions $\delta_{\operatorname{jp}}$ and $\delta_{\operatorname{cvx}}$ described in Section~\ref{sec:distance} measure how far is the set 
of discovered attributes $\mathcal{D}$ from the Meaningful Subspace $\mathcal{S}$.
The closer the distance, the more meaningful the set of attributes is.
Unfortunately, the distance may not be easy to interpret as the relationship between
the proposed distances and meaningfulness could be non-linear.
Furthermore, it is not clear how one could compare the 
results from $\delta_{\operatorname{cvx}}$ and $\delta_{\operatorname{jp}}$.

We wish to have a metric that is both easy to interpret and enables 
comparisons between various distance functions.
To that end, we consider a set of subspaces 
generated from the subspace interpolation between Meaningful Subspace and Non-Meaningful Subspace, 
or Noise Subspace.
Here, we represent Non-Meaningful Subspace $\mathcal{N}$ by a set of evenly distributed random 
attributes.

To perform the subspace interpolation, we first divide the meaningful attribute set $\mathcal{S}$ 
into two disjoint subsets $\mathcal{S}^1 \cup \mathcal{S}^2 = \mathcal{S}$.
Here, we consider the set $\mathcal{S}^1$ as the representation of the Meaningful Subspace.
The interpolated set of subspaces is generated by progressively adding random attributes,
$\tilde{\mathcal{N}} \in \mathcal{N}$ into $\mathcal{S}^2$.
The following proposition guarantees that the interpolation generates subspaces between the 
Meaningful Subspace and the Non-meaningful Subspace.
\begin{proposition}
	Let $\tilde{\mathcal{S}} = \mathcal{S}^2 \cup \tilde{\mathcal{N}}$; when $\tilde{\mathcal{N}} = \{ \}$, then the distance $\delta^*$ between 
	$\tilde{\mathcal{S}}$ and $\mathcal{S}^1$ is minimized. However, when 
	$\tilde{\mathcal{N}} \rightarrow \mathcal{N}$, then the distance between $\tilde{\mathcal{S}}$ and $\mathcal{S}^1$ is asymptotically close to $\delta^*(\mathcal{S}^1,\mathcal{N}; \mathcal{X})$, where $\delta^*$ is the distance function presented previously such as $\delta_{\operatorname{jp}}$ and $\delta_{\operatorname{cvx}}$. More precisely, we can define the relationship as follows:
	\begin{equation}
	\lim_{|\tilde{\mathcal{N}}| \rightarrow \infty} \delta^*(\tilde{\mathcal{S}},\mathcal{S}^1; \mathcal{X}) = \delta^*(\mathcal{N},\mathcal{S}^1; \mathcal{X}).
	\end{equation}
\end{proposition}

\noindent
\textit{Remarks.}
The above proposition basically says that initially when random attributes are not added into $\tilde{\mathcal{S}}$, the subspace will be close to the Meaningful Subspace $\mathcal{S}^1$. 
Furthermore, if we progressively add random attributes into $\tilde{\mathcal{S}}$, 
eventually the $\tilde{\mathcal{S}}$ will occupy a subspace that is asymptotically close to the Noise Subspace.
While it is easy to prove the above Proposition, we present one version of the proof in the supplementary material~\footnote{This material will also be available in a permanent web page once the paper has been published}. 
%By using the above Proposition, we can enumerate the points along the geodesic by progressively adding random attributes to $\mathcal{S}^2$.

%We note our subspace interpolation method is in a similar spirit to more complex methods ~\cite{gopalan2014unsupervised, gopalan2011domain} 
%that use Grassmann manifold to perform subspace analysis.
%Here, each subspace is considered as a point over a Grassmann manifold.
%Subspace interpolation is merely enumerating points along geodesic curve between two subspaces.%
%$We note that the interpolation can be thought as a geodesic curve between those two subspaces on a subspace manifold; a similar technique used in~%\cite{gopalan2014unsupervised, gopalan2011domain} that uses Grassmann manifold to perform subspace analysis.
%Unlike the manifold $\mathcal{M}$ introduced earlier, each element in the manifold of subspace is represented as a set of attributes. 
%More precisely, a set of attributes such as $\mathcal{D}$ or $\mathcal{S}$ is considered as a point in this manifold. 

%Let $\delta^{\mathcal{D}}$ is the distance of the set discovered attribute $\mathcal{D}$ to the Meaningful Subspace.
%Once we enumerate the points, then we find the closest
%point on the geodesic that has distance to the Meaningful Subspace similar to $\delta^{\mathcal{D}}$.

Let $\delta^{\tilde{\mathcal{S}}}$ be the distance between $\tilde{\mathcal{S}}$ and the Meaningful Subspace $\mathcal{S}^1$ and
$\delta^{{\mathcal{D}}}$ be the distance between $D$ and the Meaningful Subspace $\mathcal{S}^1$.
After the interpolated subspaces have been generated, we find the subspace $\tilde{\mathcal{S}}$ that makes $\delta^{\tilde{\mathcal{S}}} \approx \delta^{{\mathcal{D}}}$.
Our idea is that if $\delta^{\tilde{\mathcal{S}}} \approx \delta^{{\mathcal{D}}}$, then the meaningfulness between $\tilde{\mathcal{S}}$ and $\mathcal{D}$ should be similar.
Since, we define $\tilde{\mathcal{S}}$ as a set of meaningful attributes added with additional noise attributes, therefore, we can use this description to describe the meaningfulness of $\mathcal{D}$.
We can define this task as an optimization problem as follows:
\begin{equation}
\label{eq:calibration}
g^* = \mathop{ \arg \min_{|\tilde{\mathcal{N}}|} \left\| {\delta^*(\{\mathcal{S}^2 \cup \tilde{\mathcal{N}\}},{\mathcal{S}^1}; \mathcal{X}) - \delta^*(\mathcal{D},{\mathcal{S}^1}; \mathcal{X})} \right\|_2^2}.
\end{equation}
\noindent
where $g^*$ is the minimum number of random attributes required to be added into $\tilde{\mathcal{S}}$ so that $\delta^{\tilde{\mathcal{S}}} \approx \delta^{{\mathcal{D}}}$. 
The above optimization problem can be thought as finding the furthest subspace $\tilde{\mathcal{S}}$ from the Meaningful Subspace in an open ball 
of radius $\delta^{{\mathcal{D}}}$.
We can simply solve the above equation using
a curve fitting approach. 
In our implementation we use the least square approach.

Finally, our proposed attribute meaningfulness metric, $\gamma$ is defined as follows.
\begin{equation}
\label{eq:calibration}
\gamma(\mathcal{D};\mathcal{X}, \mathcal{S})=\left({1 -{{g^*} \over {|{\mathcal{S}^2}|} + g^*}}\right) \times 100.
\end{equation}
\noindent
\textit{Remarks.}
The proposed metric indicates how much noise/non-meaningful attributes is required to have similar distance to $\delta^{\mathcal{D}}$.
On the other hand, the metric reflects how many meaningful attributes are contained in the attribute set.
Less noise implies a more meaningful attribute set.

As different distance functions may capture different aspects of meaningfulness, it is possible to combine them under the proposed metric.
In our work, we use a simple equally weighted summation, 
$\tilde{\gamma} = \frac{1}{2} \gamma_{\operatorname{cvx}} + \frac{1}{2} \gamma_{\operatorname{jp}}$, as our final metric.

%More precisely, a random attribute $\Vec{v}_i \in \mathbb{R}^N$ is generated by first generating random vector from Gaussian distribution with
%zero mean and unit variant.
%Then, we threshold the value to by applying $\operatorname{sign}$ function to the vector.
%The $\operatorname{sign}(t)$ function gives 1 when the $t > 0$, and 0 otherwise.

%Since there is no prior assumption to this manifold  structure, then it is not possible
%to analytically derive the geodesic equation.
%Henceforth, we resort to an approximation method by enumerating points in along the geodesic.

\section{Experiments}
\label{Experiment}

In this section, we first evaluate the ability of our approach to measure the meaningfulness of a set of attributes.
Then, we use our proposed metric to evaluate attribute meaningfulness on
the attribute sets generated from various automatic attribute discovery methods such as
PiCoDeS~\cite{bergamo2011picodes} and
Discriminative Binary Codes~(DBC)~\cite{rastegari2012attribute} as well as
the hashing methods such as
Iterative Quantization (ITQ)~\cite{gong2011iterative},
Spectral Hashing (SPH)~\cite{weiss2009spectral} and
Locality Sensitivity Hashing (LSH)~\cite{Leskovec2009Mining}.
Then we perform a user study on two datasets to validate the effectiveness of the proposed metric. 

We apply the two metrics $\gamma_{\operatorname{jp}}$
~\eqref{eq:joint_prob}, $\gamma_{\operatorname{cvx}}$ ~\eqref{eq:con_hull} and the combined metric $\tilde{\gamma}$ to
compare the meaningfulness of the attributes discovered by the above methods
on three attribute datasets: (1) a-Pascal a-Yahoo dataset
(ApAy)~\cite{Farhadi09describingobjects}; (2) Animal with
Attributes dataset (AwA)~\cite{Lampert13} and; (3) SUN Attribute
dataset (ASUN)~\cite{Patterson2012SunAttributes}.

A user study is performed to test the meaningfulness of attributes discovered by each method on ApAy and ASUN datasets. 

%Then we will respectively evaluate and discuss the results of transformation framework and results of log likelihood.
%Finally, we will show some qualitative results of the attributes discovered by the comparative attribute discovery methods.
%This will render a intuitive perception to the readers and show our measurement score is directly related to the extent of the visual semanticness that can be understandable by human.

\subsection{Datasets and experiment setup}
\noindent
\textbf{a-Pascal a-Yahoo dataset (ApAy)~\cite{Farhadi09describingobjects} --- } comprises two sources: a-Pascal and a-Yahoo.
There are 12,695 cropped images in a-Pascal that are divided into
6,340 for training and 6,355 for testing with 20 categories.
The a-Yahoo set has 12 categories disjoint from the a-Pascal categories.
Moreover, it only has 2,644 test exemplars.
There are 64 attributes provided for each cropped image.
The dataset provides four features for each exemplar: local texture; HOG; edge and color descriptor.
We use the training set for discovering attributes and we perform our study on the test set. More precisely, we consider the test set as the set of images $\mathcal{X}$.

\noindent
\textbf{Animal with Attributes dataset (AwA)~\cite{Lampert13} --- } the dataset contains 35,474 images of 50 animal categories with 85 attribute labels.
There are six features provided in this dataset: HSV color histogram; SIFT~\cite{lowe2004distinctive}; rgSIFT~\cite{van2010evaluating}; PHOG~\cite{bosch2007representing}; SURF~\cite{bay2008speeded} and local self-similarity~\cite{shechtman2007matching}.
AwA dataset is proposed for studying the zero-shot learning problem.
As such, the training and test categories are disjoint; there are no training images for test categories and vice versa.
More specifically, the dataset contains 40 training categories and 10 test categories.
Similar to the ApAy dataset, we use the training set for discovering attributes and we perform the study in the test set.

\noindent
\textbf{SUN Attribute dataset (ASUN)~\cite{Patterson2012SunAttributes} --- } ASUN is a fine-grained scene classification dataset
consisting of 717 categories (20 images per category) and 14,340 images in total with 102 attributes.
There are four types  of features provided in this dataset: (1) GIST; (2) HOG;
(3) self-similarity and (4) geometric context color histograms~(See \cite{xiao2010sun} for feature and kernel details).
From 717 categories, we randomly select 144 categories for discovering attributes.
As for our evaluation, we random select 1,434 images (\ie 10\% of 14,340 images) from the dataset.
It means, in our evaluation, some images may or may not come from the 144 categories used for discovering attributes.

For each experiment, we apply the following pre-processing described in~\cite{bergamo2011picodes,YangMNCH15}.
We first lift each feature 
into a higher-dimensional
%space approximating the histogram intersection kernel by using the
%explicit feature maps proposed by Vedaldi and Zisserman~\cite{Vedaldi2012Efficient}.
%More precisely, each feature is mapped 
into the space three times larger than the original
space.
%This effectively allows us to apply linear classifiers in the explicit kernel space~\cite{bergamo2011picodes}.
After the features are lifted, we then apply PCA to reduce the dimensionality of the feature space by 40 percent.
This pre-processing step is crucial for PiCoDeS as it uses lifted feature space to simplify their training scheme while maintaining the information preserved in the Reproducing Kernel Hilbert Space~(RKHS).
Therefore, the method performance will be severely affected when lifting features are not used.
%In our empirical observations (results not presented), we also found that lifted feature space gives positive contributions to the other methods.

%improves other methods performance.

%Incorporating the preprocessing step is because some approaches such as PiCoDeS use lifted feature space to simplify their training scheme while maintaining the information preserved in the Reproducing Kernel Hilbert Space~(RKHS). So these methods will be severely affected when lifting features are not used.

Each method is trained using the training images to discover the attributes.
Then we use the manifold $\mathcal{M}$ w.r.t. the test images for the evaluation.
More precisely, each attribute descriptor is extracted from test images (\ie $\Vec{z}_k, \Vec{z}_k \in \{-1,1\}^N$, where $N$ is the number of test images).
For each dataset, we use the attribute labels from AMT to represent the Meaningful Subspace, $\mathcal{S}$.

\begin{figure*}
	\centering
	\includegraphics[width=0.85\linewidth]{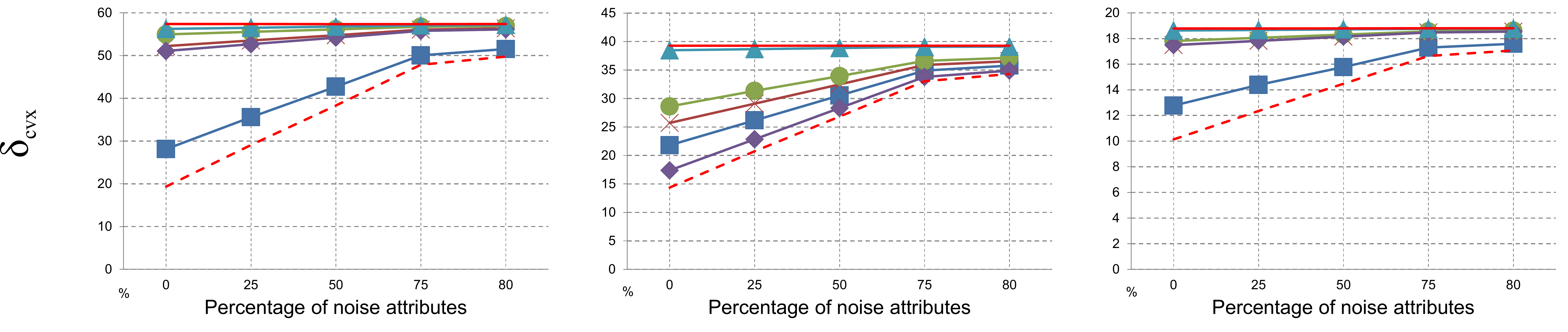}  \\
	\includegraphics[width=0.85\linewidth]{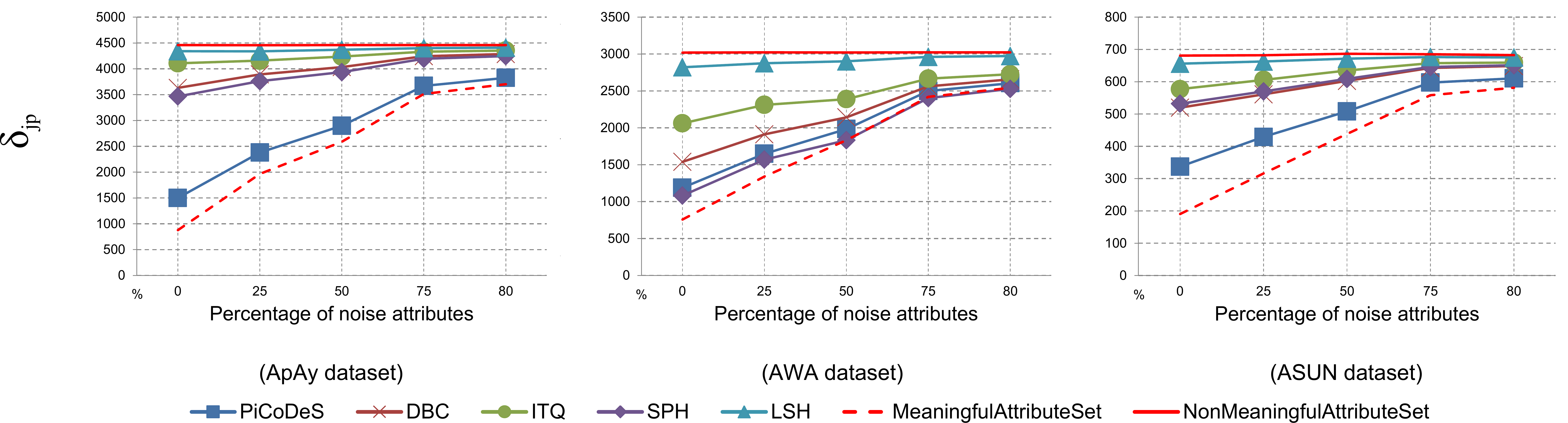}
	\caption{Validation of attribute meaningfulness measurement by reconstruction error $\delta_{\operatorname{cvx}}$ (first row) and $\delta_{\operatorname{jp}}$ (second row).
		As we can see, both distances become larger when more random/non-meaningful attributes are added.
		MeaningfulAttributeSet has the closest distance to the Meaningful Subspace and NonMeaningfulAttributeSet always has the largest distance. Here, each method is configured to discover 32 attributes. The results for different number of attributes are presented in the supplementary materials.
		The smaller the $\delta$, the more meaningfulness.
	}
	\label{fig:noise}
\end{figure*}

\subsection{Do $\delta_{\operatorname{cvx}}$ and $\delta_{\operatorname{jp}}$ measure meaningfulness?}
%\subsection{Validation of Attribute Meaningfulness by Reconstruction Error}
% adding noise experiment
In this experiment, our aim is to verify whether the proposed
approach does measure meaningfulness on the set of discovered
attributes. 
One of the key assumptions in our proposal is that the meaningfulness is reflected from the distance between the Meaningful Subspace and the given attribute set, $\mathcal{D}$.
That is, if the
distance is far, then it is assumed that the attribute set is less
meaningful, and vice versa.
In order to evaluate that, we create two sets of attributes,
meaningful and non-meaningful attributes, and observe their
distances to the meaningful subspace.

%Our work also depends on the speculation posed in Parikh and Grauman's work that Meaningful Subspace has a \textit{shared structure}.
%Hence, the evaluation results in this part also gives indication whether this speculation is true (\ie whether the \textit{shared structure} does exist).

%A big question in our work up to now is 'Are these two approaches $\delta_{\operatorname{jp}}$ and $\delta_{\operatorname{cvx}}$ really able to measure the meaningfulness of a set automatic discovered attributes'.

For the meaningful attribute set, we use the attributes from AMT
provided in each dataset. More precisely, given manually labelled attribute set $\mathcal{S}$, we divide the set into two subsets $\mathcal{S}^1
\cup \mathcal{S}^2 = \mathcal{S}$. Following the method used in
Section~\ref{sec:part3}, we use $\mathcal{S}^1$ to represent the
Meaningful Subspace and consider $\mathcal{S}^2$ as a set of
discovered attributes (\ie $\mathcal{D} = \mathcal{S}^2$). As
human annotators are used to discover $\mathcal{S}^2$, these attributes are
considered to be meaningful. We name this as the
\textit{MeaningfulAttributeSet}.

For the latter, we generate attributes that are not meaningful by
random generation. More precisely, we generate a finite set of random
attributes $\tilde{\mathcal{N}}$ following the method described in
Section~\ref{sec:part3}. As the set $\tilde{\mathcal{N}}$ is
non-meaningful, it should have significantly larger distance to the Meaningful
Subspace. We name this set as \textit{NonMeaningfulAttributeSet}.
Furthermore, we progressively add random attributes to the set of
attributes discovered from each method, to evaluate whether the distance to Meaningful Subspace is
enlarged when the number of non-meaningful attributes increases.
%(\ie $\tilde{\mathcal{D}} = \Vec{z} \cup \mathcal{D}, \Vec{z} \in \mathcal{N}$).

Fig.~\ref{fig:noise} presents the evaluation results. Due to space
limitation, we only present the results of the case where 32 attributes
are discovered by the methods. We present the
remaining results in the supplementary materials. From the results, it
is clear that \textit{MeaningfulAttributeSet} has the closest
distance to the Meaningful Subspace in all datasets for both
distances $\delta_{\operatorname{cvx}}$ and
$\delta_{\operatorname{jp}}$. As expected the
\textit{NonMeaningfulAttributeSet} has the largest distance
compared with the others. In addition, as more random attributes
are added, the distance between the sets of attributes discovered
for every approach and the Meaningful Subspace increases. These
results indicate that the proposed approach could measure the set
of attribute meaningfulness. In addition, these also give a strong
indication that meaningful attributes have
\textit{shared structure}.

\begin{figure*}
	\centering
	\includegraphics[width=0.85\linewidth]{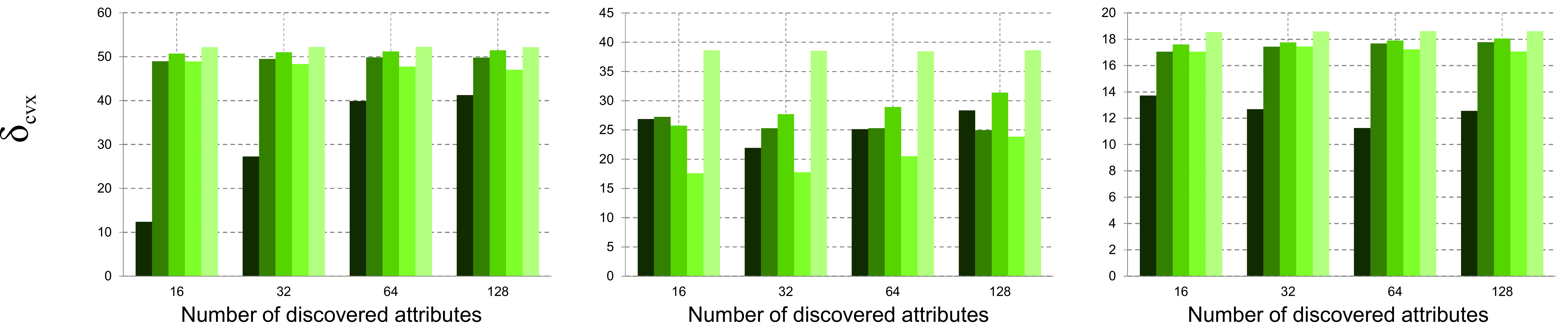}  \\
	\includegraphics[width=0.85\linewidth]{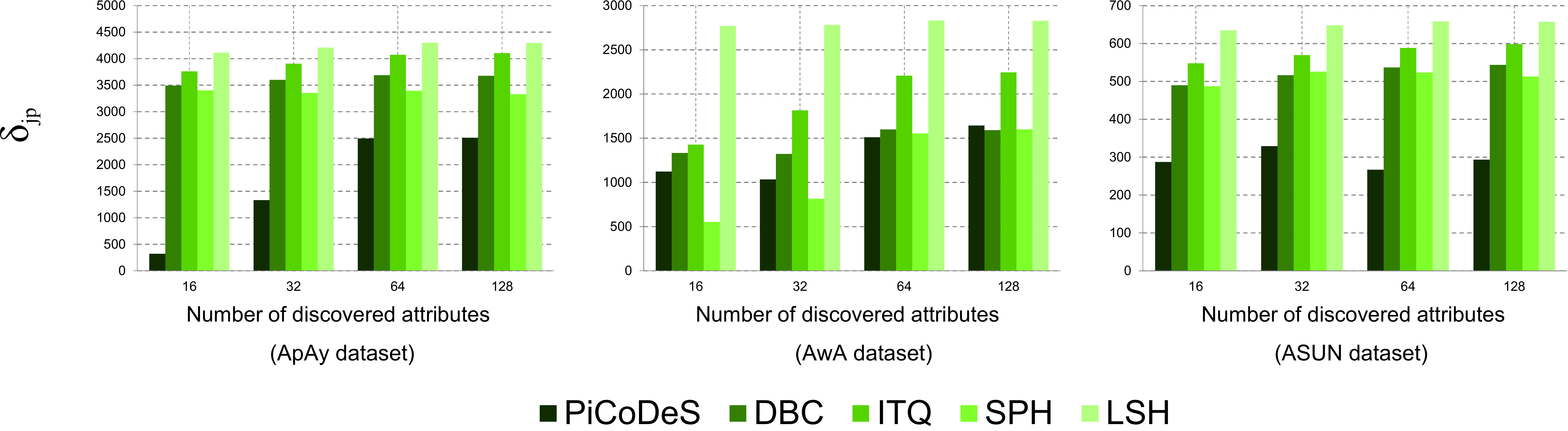}
	\caption{\textit{Attribute meaningfulness} comparisons between different methods on variant number of discovered attributes. The first row reports the results using $\delta_{\operatorname{cvx}}$ and the second row reports the results using $\delta_{\operatorname{jp}}$.
		The smaller the $\delta$, the more meaningfulness.}
	\label{fig:recon}
\end{figure*}

%\begin{tabular}{c}
%    \centering
%    \includegraphics[width=0.8\linewidth]{compare_result_convex_hull_trim.pdf}  \\
%    \hline
%    \includegraphics[width=0.8\linewidth]{compare_result_joint_prob_comp__jpgd2recon_trim.pdf}
%        \caption{\textit{Attribute meaningfulness} comparisons between different number of discovered attributes. The first row reports the results using $\delta_{\operatorname{cvx}}$ and the second row reports the results using $\delta_{\operatorname{jp}}$.}
%             \label{fig:recon}
%\end{tabular}

%\subsection{Attribute Meaningfulness Evaluation using $\delta_{\operatorname{cvx}}$ and $\delta_{\operatorname{jp}}$}

\subsection{Attribute set meaningfulness evaluation using $\delta_{\operatorname{cvx}}$ and $\delta_{\operatorname{jp}}$}
\label{sec:att-eval}
% origninal experiment
In this section, we evaluate the meaningfulness for the set of
attributes automatically discovered by various approaches in the
literature. To that end, for each dataset, we use all of the sets of
attributes from AMT as the representation of the Meaningful Subspace.
Then, we configure each approach to discover 16, 32, 64 and 128 attributes.

Fig.~\ref{fig:recon} reports the evaluation results in all
datasets. It is important to point out that as the distance is
not scaled, we can only analyse the results in terms of rank
ordering (\ie which method is the best and which one comes the second).

PiCoDes has the lowest distance on most of the datasets with variant number of attributes extracted. It uses category labels and applies max-margin framework to jointly learn the category classifier and attribute descriptor in an attempt to maximizing the descriptor discriminative power. In other words, PiCodes is aimed to discover a set of attributes that could discriminate between categories.

DBC also use maximum-margin technique to extract meaningful attributes. 
However, DBC discovers less meaningful attributes than PiCoDeS.
We conjecture that this could be due to the fact that unlike PiCoDeS that learns the
attribute individually, DBC learns the whole attribute descriptor for each category simultaneously.
This scheme will inevitably put more emphasis on category discriminatebility of attribute rather than preserving the individual attribute meaningfulness.
Note that here we do not suggest that DBC does not discover meaningful attributes,
rather, PiCoDeS may find more meaningful attributes.
Therefore, our finding does not contradict the results presented in the DBC original paper \cite{rastegari2012attribute} suggesting that
the method does find meaningful attributes.

Another observation is that the results indicate that SPH discover meaningful attributes.
SPH aims to find binary codes by preserving the local neighborhood structure via a graph embedding approach.
%In fact, it is the best method in the AwA dataset.
One possible explanation could be that when two images belong to the same category,
they should share more attributes indicating a shorter distance
between them in the binary space, and vice versa.

Despite its goal to learn similarity preserving binary descriptor,
ITQ has a larger distance than SPH, DBC and PiCoDeS.
ITQ learns the binary descriptor by using the global information of the data distribution. More precisely, it minimizes the quantization error of the mapping data to the vertices of a zero centered binary hypercube. This suggests that using only global information might not be effective to discover meaningful attributes.

It is expected that LSH has the highest distance to the Meaningful Subspace (\ie least meaningfulness).
LSH uses random hyperplanes to project a data point into the binary
space.
Thus, there are no consistent identifiable visual concepts
presented in the positive images.

In summary, these results suggest two recipes that could be
important in developing attribute discovery methods: the method
should attempt to discover discriminative attributes as well as to
preserve local neighborhood structure.

\subsection{Attribute set meaningfulness calibration using the proposed meaningfulness metric}
As shown in section~\ref{sec:att-eval}, the distance between
attribute sets and the meaningful subspace is uncalibrated, which makes it hard to
quantitatively compare different methods. The proposed meaningfulness
metric will convert distances to scores and enable the
quantitative analysis.
\noindent
\begin{figure}[htb]
	\centering
	\includegraphics[width=0.8\linewidth]{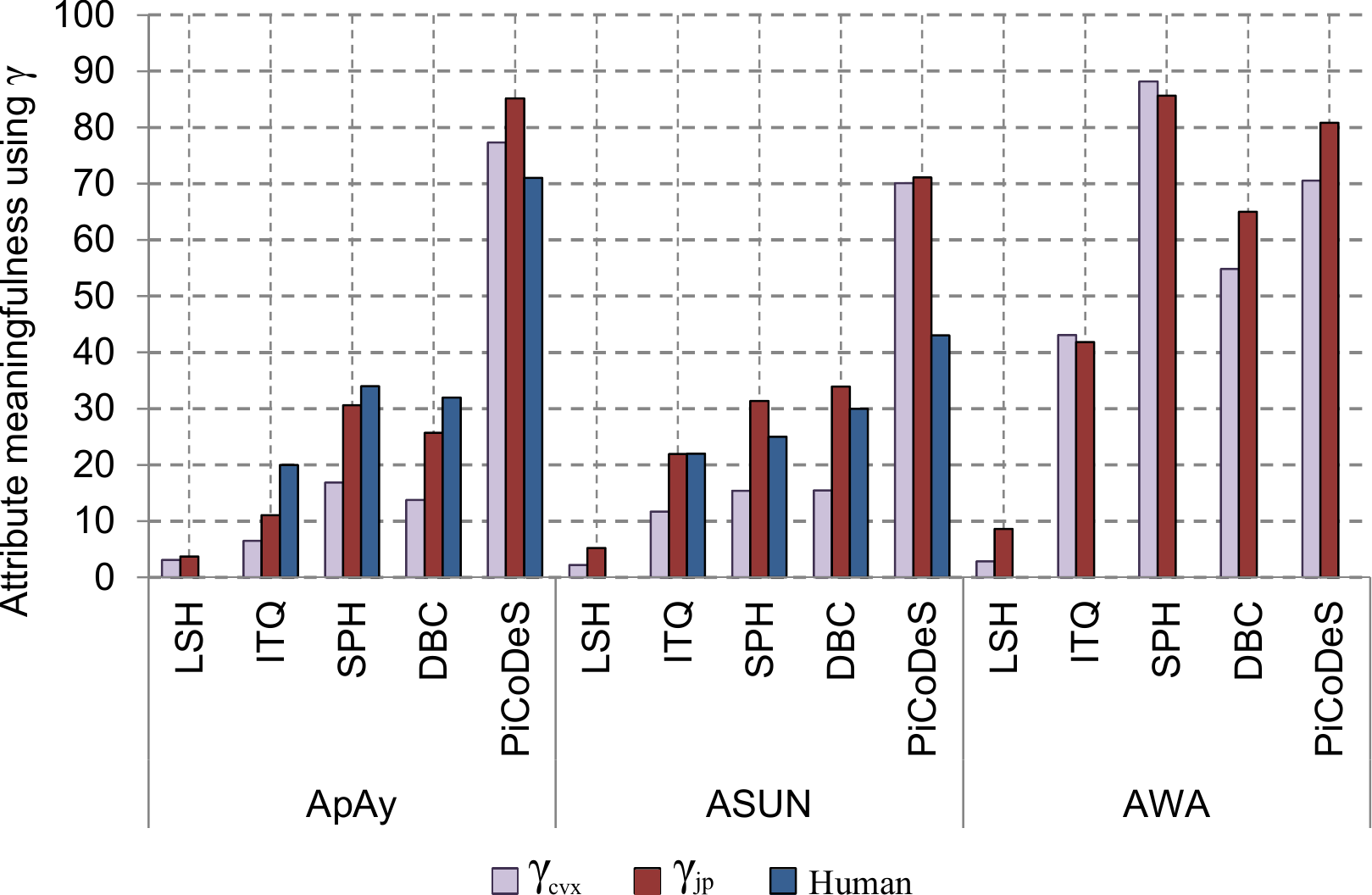}
	\caption{Comparisons of various methods using the proposed meaningfulness metric as well as human study results. 
		Each method is set to discover 32 attributes. 
		The higher the more meaningful. 
		Human study is not conducted for AWA dataset as special zoology knowledge is required.
		The human results for LSH method are 0 for ApAy and ASUN datasets.
	}
	\label{fig:calibration}
\end{figure}

Now we apply the metrics $\gamma_{\operatorname{cvx}}$ and $\gamma_{\operatorname{jp}}$ by calibrating the distances $\delta_{\operatorname{cvx}}$ and $\delta_{\operatorname{jp}}$ as shown in (\ref{eq:calibration}). 
Fig.~\ref{fig:calibration} presents the results when the methods are configured to discover 32 attributes. 
The ranking orders of five methods according to $\gamma_{\operatorname{cvx}}$ and $\gamma_{\operatorname{jp}}$ are the same with similar values in most tests, with two exceptions in ASUN dataset. 
One possible reason is that each metric captures a
different aspect of the attribute meaningfulness. 
The proposed
$\gamma_{\operatorname{cvx}}$ captures a one-to-many relationship and $\gamma_{\operatorname{jp}}$
evaluates the one-to-one relationship. We then use the equal weighted metric score $\tilde{\gamma}$ for further analysis.

We also perform a user study on the outcome attributes from each attribute discovery methods. 
We only use ApAy and ASUN dataset for user study, since AwA requires experts in animal studies.
The study collected over 100 responses. Each response presents positive and negative images of 8 discovered attributes randomly selected. 
The user was asked whether these two set of images represent a consistent visual concept (hence meaningful).
The responses were averaged by considering 1 as meaningful and 0 as non-meaningful.
These users were the staff and students with different background knowledge from various major fields in the University including IT, Electronic Eng., History, Philosophy, Religion and Classics and Chemical Eng.

Table~\ref{tab:noise} shows both the the results of $\tilde{\gamma}$ and the human study.
Again, we see that the attribute set discovered by LSH has the
lowest meaningful content close to 0\%. Thus, LSH generates the
least meaningful attribute sets. PiCoDeS and SPH generally
discover meaningful attribute sets with much less noise. The
randomized methods such as LSH and ITQ tend to generate less
meaningful attribute sets with attribute meaningfulness around 1\%-20\%. By applying learning techniques such as PiCodes, DBC and SPH, the attribute meaningfulness could be significantly increased (\ie on average by 10-20 percentage points). % to 15\%-80\%.

The results on the user study show similar trends as the results from the proposed metric.
In addition, we also compare the user study results with $\gamma_{\operatorname{cvx}}$ and $\gamma_{\operatorname{jp}}$ in Fig.~\ref{fig:calibration}. 
The trend is still consistent.

\begin{figure}[htb]
	\centering
	\includegraphics[width=0.7\linewidth]{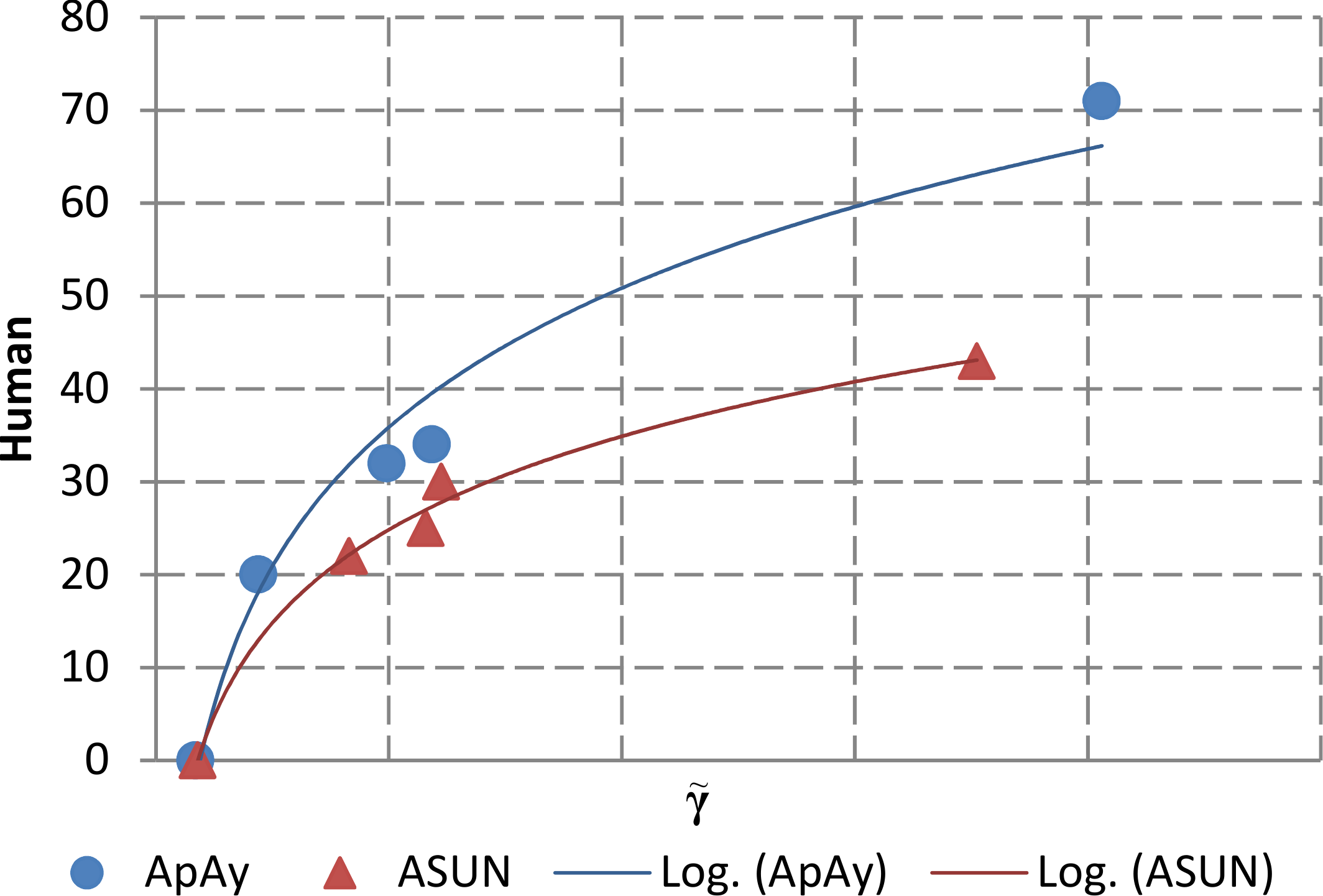}
	\caption{The demonstration of correlation between the results of $\tilde{\gamma}$ and user study on both ApAy and Asun datasets.}
	\label{fig:calibration_plot}
\end{figure}

The correlation of the user study results and the score of the metric is shown in Fig.~\ref{fig:calibration_plot} by applying a simple logarithmic fitting using the data from Table~\ref{tab:noise}. 
This demonstration indicates that our method is, to some extend, able to measure the meaningfulness of a set
of discovered attributes as human does via a simple non-linear regression.

It is noteworthy to mention that the time cost of the evaluation by our metric is significantly smaller than the manual process using AMT.
Recall that, the time required for a human annotator (an AMT worker) to finish one HIT is 2 minutes, an AMT worker may need 320 minutes to finish evaluating  5 methods wherein each is configured to discover 32 attributes.
Our approach only needs 105 seconds in total to evaluate the whole three datasets (\ie 35 seconds each);
thus, leading to several orders of magnitude speedup!

%\captionsetup[table]{labelfont={color=red}}
\begin{table}[htbp]
	\centering
	\small
	\caption{The results of meaningfulness metric $\tilde{\gamma}$ on the three datasets and the results (in percentage of meaningfulness) of user study on ApAy \& ASUN when each method is configured to discover 32 attributes.
		The bold text indicates the top performing method in the proposed metric.
		The higher the more meaningful.}
	\scalebox{0.8}{
		\begin{tabular}{c|cc|cc|cc}
			\toprule
			\parbox[t]{1.2cm}{Methods\\\textbackslash Datasets} & \multicolumn{2}{c|}{ApAy} & \multicolumn{2}{c|}{ASUN} & \multicolumn{2}{c}{AwA} \\
			\midrule
			& $\tilde{\gamma}$  & Human & $\tilde{\gamma}$  & Human & $\tilde{\gamma}$ & Human \\
			\cline{2-7}     
			LSH   & 3.4   & 0    & 3.6   & 0   & 5.9 & \multirow{5}{*}{N/A} \\
			ITQ   & 8.8   & 20   & 16.6  & 22  & 42.6 &\\
			SPH   & 23.7  & 34   & 23.2  & 25  & \textbf{86.9} & \\
			DBC   & 19.8  & 32   & 24.5  & 30  & 60 &\\
			PiCoDeS & \textbf{81.2} & \textbf{71}  & \textbf{70.5} & \textbf{43}    & 75.7 &\\
			\bottomrule
		\end{tabular}%
		
		\label{tab:noise}%
	}
\end{table}%

\section{Conclusions}
\label{sec_conclusion}

In this paper, we studied a novel problem of measuring the meaningfulness of automatically discovered attribute sets.
To that end, we proposed a novel metric, here called the \textit{attribute meaningfulness} metric.
We developed
two distance functions for measuring the meaningfulness of a set of attributes.
The distances were then calibrated by using subspace interpolation between Meaningful Subspace and Non-meaningful/Noise Subspace.
%This interpolation can be considered as a geodesic between two subspace in a subspace manifold.
The final metric score indicates how much meaningful content is contained within the set of discovered attributes.
In the experiment, the proposed metrics were used to evaluate the \textit{meaningfulness} of attributes discovered by two recent automatic attribute discovery methods and three hashing methods on three datasets.
A user study on two datasets showed that the proposed metric has strong correlation to human responses.
The results concluded that there is a strong indication that the shared structure could exist among the meaningful attributes.
The results also give evidence that discovering attributes by optimising the attribute descriptor discrimination and/or 
preserving the local similarity structure 
could yield more meaningful attributes.
In future work, we plan to explore other constraints or optimisation models to capture the hierarchical structure of semantic concepts.
Moreover, the semantic concepts discovery in complex uncontrolled long video~\cite{ChangYXY15a} could also be a good scenario to extend our proposed metric.
%Moreover, the video semantic analysis~\cite{ChangYLZH16,ChangYHXY15} is also an appropriate scenario to apply the proposed metric for meaningfulness evaluation of video-base visual attribute.
Some other directions could be to investigate the influence of the degenerated or low-resolution image~\cite{liu2012novel} on the attribute meaningfulness evaluation or to evaluate the potential attributes for 3D reconstructed image sequences~\cite{Zhu_2014_CVPR}.
We also plan to perform more large-scale user studies on AMT on other datasets.

{\small
\bibliographystyle{ieee}
\bibliography{egbib}
}

\end{document}